\documentclass[letterpaper, 10 pt, conference]{ieeeconf}  

\IEEEoverridecommandlockouts                              

\usepackage{graphicx}
\usepackage{authblk}
\usepackage{amsmath}
\usepackage{amssymb}
\usepackage{booktabs}
\usepackage{color,soul}
\usepackage[utf8]{inputenc}
\usepackage{standalone}
\usepackage{float}
\usepackage{multirow}
\usepackage{caption}
\usepackage{url}
\makeatletter
\@namedef{ver@everyshi.sty}{}
\makeatother
\usepackage{tikz}
\usepackage{todonotes}
 
\usetikzlibrary{decorations.pathreplacing,calligraphy}
\usepackage{pgfplots}
\let\labelindent\relax
\usepackage{enumitem}
\usepackage[capitalize]{cleveref}
\crefname{section}{Sec.}{Secs.}
\Crefname{section}{Section}{Sections}
\Crefname{table}{Table}{Tables}
\crefname{table}{Tab.}{Tabs.}

\title{\LARGE \bf
RACECAR - The Dataset for High-Speed Autonomous Racing
}

\author{%
Amar Kulkarni$^{1}$, \quad John Chrosniak$^{1}$, \quad Emory Ducote$^1$, \quad Florian Sauerbeck$^2$, \quad Andrew Saba$^3$, \\
Utkarsh Chirimar$^1$ \quad John Link$^1$, \quad Marcello Cellina$^4$, \quad Madhur Behl$^1$\\
$^1$University of Virginia,$^2$Technical University of Munich,$^3$Carnegie Mellon University,$^4$Politecnico di Milano\\
\texttt{\{ark8su,jlc9wr,etd4sv,uc6gq,jwl9vq,madhur.behl\}@virginia.edu}\\
\texttt{\{florian.sauerbeck\}@tum.de\,\{asaba\}@andrew.cmu.edu\,\{marcello.cellina\}@polimi.it}
}

\begin{document}

\twocolumn[{%
\renewcommand\twocolumn[1][]{#1}%
\maketitle
\begin{center}
    \centering
    \captionsetup{type=figure}
    \includegraphics[width=\linewidth]{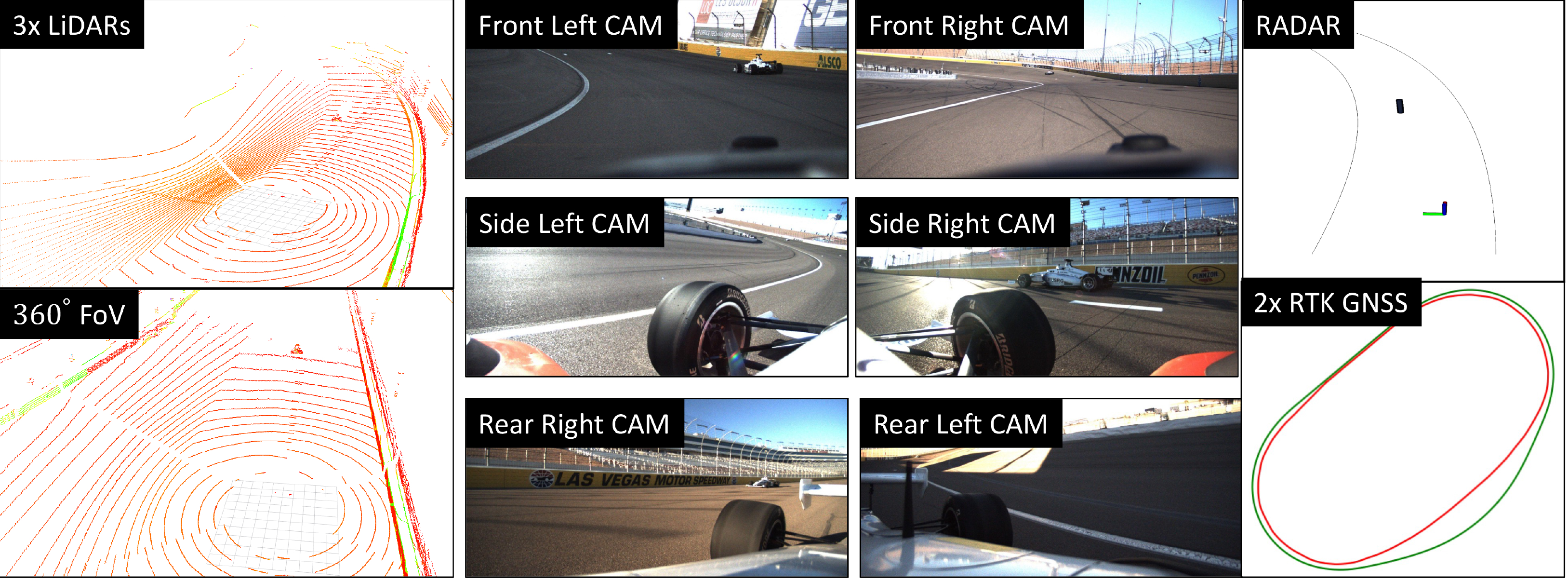}
    \captionof{figure}{RACECAR is the first multi-model sensor data collected from fully autonomous Indy race cars operating at speeds of up to 170 mph (274 kph). The dataset spans 11 racing scenarios with over 6.5 hours of track activity.}
    \label{fig:overview}
\end{center}%
}]

\thispagestyle{empty}
\pagestyle{empty}

\begin{abstract}

This paper describes the first open dataset for full-scale and high-speed autonomous racing.
Multi-modal sensor data has been collected from fully autonomous Indy race cars operating at speeds of up to 170 mph (273 kph).
Six teams who raced in the Indy Autonomous Challenge have contributed to this dataset.
The dataset spans 11 interesting racing scenarios across two race tracks which include solo laps, multi-agent laps, overtaking situations, high-accelerations, banked tracks, obstacle avoidance, pit entry and exit at different speeds.
The dataset contains data from 27 racing sessions across the 11 scenarios with over 6.5 hours of sensor data recorded from the track.
The data is organized and released in both ROS2 and nuScenes format. We have also developed the ROS2-to-nuScenes conversion library to achieve this.
The RACECAR data is unique because of the high-speed environment of autonomous racing.
We present several benchmark problems on localization, object detection and tracking (LiDAR, Radar, and Camera), and mapping using the RACECAR data to explore issues that arise at the limits of operation of the vehicle. 
\end{abstract}

\section{Introduction}\label{sec:intro}
While autonomous vehicle research and development is focused on handling routine driving situations, achieving the safety benefits of autonomous vehicles also requires a focus on driving at the limits of the control of the vehicle.  
Demonstrating high-speed autonomous racing can be considered as a grand challenge for autonomous driving and making progress here has the potential to enable breakthroughs in agile and safe autonomy.
\begin{figure*}
     \centering
     \includegraphics[width=\linewidth]{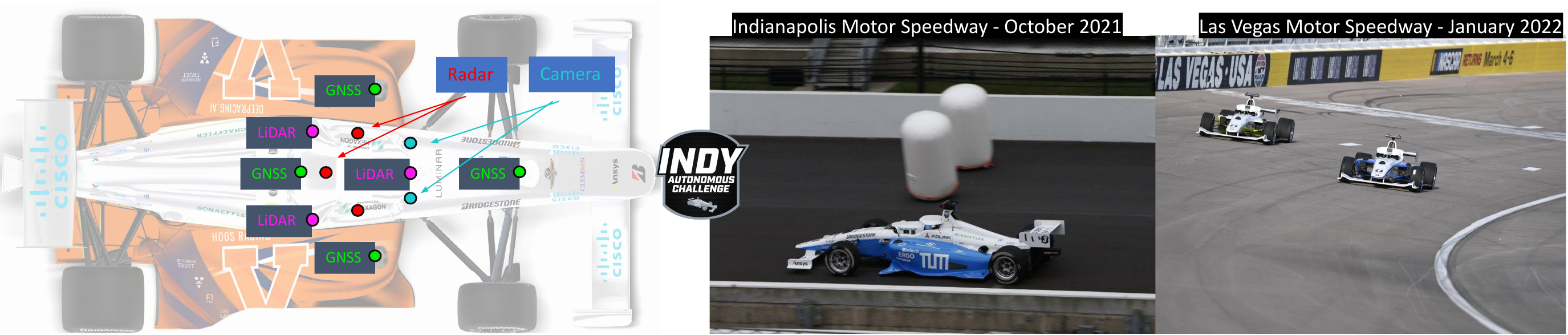}
     \caption{\textbf{[Left]} The AV-21 is a modified Indy Lights racecar retrofitted with 3 LiDARs, 6 Cameras, 3 Radars, 2 GNSS systems. \textbf{[Right]} The Indy Autonomous Challenge (IAC) held its first autonomous race at the Indianapolis Motor Speedway (IMS) track in 2021, followed by a head-to-head overtaking competition held at the Las Vegas Motor Speedway (LVMS).}
     \label{fig:iac}
     \vspace{-10pt}
 \end{figure*}
In recent years autonomous, racing competitions, such as F1/10 autonomous racing~\cite{o2019f1,babu2020f1tenth}, Indy Autonomous Challenge~\cite{indy,wischnewski2022indy,hartmann2021autonomous}, and Formula SAE Driverless~\cite{kabzan2020amz,zeilinger2017design} are becoming proving grounds for testing motion planning and control algorithms at high speeds.
While the autonomous racing community and research have grown~\cite{betz2022autonomous} by an order of magnitude due to these competitions, the field is still very specialized and exclusive.  
Full-scale autonomous racing like the Indy Autonomous Challenge requires a large research team to develop the autonomous racing stack. As such, there are only 9 university teams in the world that own and operate full-scale fully-autonomous Indy race cars. The barrier to entry into autonomous racing is high and the underlying research challenges remain elusive from the reach of the larger computer vision, machine learning, and robotics communities. 

This paper presents the RACECAR dataset which contains multi-modal sensor data collected from the Indy Autonomous Challenge during the 2021-22 racing season.  
The autonomous race cars were outfitted with a full sensor suite for localization and perception, including solid state LiDARs, high precision RTK GNSS and IMU units, multiple cameras, and radar sensors.
This paper has the following contributions:
\begin{enumerate}[noitemsep,topsep=0pt]
    \item We present the RACECAR dataset describing the unique features, and autonomous racing contexts within which this data was collected. 
    \item We developed a ROS2 to nuScenes conversion library which allows us to release the dataset in both ROS2 bag file format, ubiquitously used by the robotics community, as well as in nuScenes format. 
    \item Using the RACECAR data, we provide benchmark challenges for the research community on high-speed localization, object detection, and tracking with baselines derived from techniques deployed on real autonomous racecars.
\end{enumerate}

Our intention is to democratize the field of autonomous racing, making it accessible to researchers who do not have access to a racecar. In doing so, we hope that the RACECAR data will enable further advances in perception, planning, and control for autonomous driving at its limits by making the underlying algorithms more robust and stress tested at high speeds. 
The data and associated code are located at \url{https://github.com/linklab-uva/RACECAR_DATA}

\section{Related Work}\label{sec:relatedwork}
Although there are several autonomous racing competitions held at different scales, there is no large-scale autonomous racing dataset available. 
The Formula Student Objects in Context~\cite{fsoco_2022} data provides annotated camera and LiDAR frames with track bounds indicated by colored cones. 
More generally, several large-scale autonomous driving datasets have been released in recent years, some accompanied by open benchmark challenges. 
The most notable among these is the KITTI Dataset \cite{KittiBenchmark} 
which has led to improvements and new methods for 3D object detection, visual odometry, and Simultaneous Localization and Mapping (SLAM).
Other datasets such as the Waymo Open Dataset~\cite{waymoOD}, 
and the Lyft Level 5 Data~\cite{Lyft5} are also noteworthy due to the scale, scenario coverage, and quality of annotations. 
The nuScenes Dataset \cite{nuscenes} is another popular autonomous driving dataset that includes camera, LiDAR, Radar, GPS, and CANBus data.
Also included are annotations of semantic descriptions, vehicle attributes such as velocity and pose, and cuboid bounding boxes.

The top speed of any of the vehicles within these datasets remains limited to highway driving speed.
RACECAR data is not a replacement for existing AV datasets, but instead provides a unique setting that is not possible to capture in on-road testing. Since the AV-21 racecar has a similar set of sensors as one would find on any AV prototype, the high-speed nature of the RACECAR data makes it suitable to test the limits of perception algorithms.
\section{RACECAR: Data Collection}\label{sec:racecarsetup}

The RACECAR dataset is compiled by contributions from several teams, all of whom competed in the inaugural season of the Indy Autonomous Challenge during 2021-22. 
Nine university teams participated in two races. The first race was held at the Indianapolis Motor Speedway (IMS) track in Indiana, USA in October 2021 (Fig.~\ref{fig:iac}[Right]). This track is a 2.5 mile (4 km) oval and is home to the famous 'Indy 500' race. The second race was held at Las Vegas Motor Speedway (LVMS) in January 2022 (Fig.~\ref{fig:iac}[Right]). The track is shorter (1.5 mile) and more aggressively banked (up to 20 degrees in the turns) making it challenging to run the cars at high speeds. 
At IMS, teams reached speeds up to 150 mph on straights and 136 mph in turns, competing in solo vehicle time trials and obstacle avoidance. At LVMS, teams participated in a head-to-head overtaking competition reaching speeds in excess of 150 mph!, with the fastest overtake taking place at 170 mph.


\subsection{Sensor Configuration}
The AV-21 Indy Lights vehicle (Fig.~\ref{fig:iac}[Left]) is outfitted with three radars, six pinhole cameras, and three solid-state LiDARs. Each of the sensor modalities covers a 360-degree field of view around the vehicle. For localization, the vehicle is equipped with two sets of high-precision Real-Time Kinematic (RTK) GNSS receivers and IMU.
The chassis, as well as the steering, powertrain, and brake system, are as close as possible to the base Indy Lights race car but were controlled autonomously using a custom drive-by-wire system. The top speed of the vehicle is rated at 180 mph. Detailed sensor configuration, specification, and calibration information is available

\begin{table}
  \centering
  \begin{tabular}{||l l l l||}
    \hline
    \textbf{Scenario} & \textbf{Track} & \textbf{Description} & \textbf{Speeds} \\
    \hline
    $S_{1}$ & LVMS & Solo Slow Lap & $<$ 70 mph \\
    $S_{2}$ & LVMS & Solo Slow Lap & 70-100 mph \\
    $S_{3}$ & LVMS & Solo Fast Lap & 100-140 mph\\
    $S_{4}$ & LVMS & Solo Fast Lap & $>$ 140 mph\\
    $S_{5}$ & LVMS & Multi-Agent Slow & $<$ 100 mph\\
    $S_{6}$ & LVMS & Multi-Agent Fast & $>$ 130 mph\\
    $S_{7}$ & IMS & Solo Slow Lap & $<$ 70 mph \\
    $S_{8}$ & IMS & Solo Slow Lap & 70-100 mph \\
    $S_{9}$ & IMS & Solo Fast Lap & 100-140 mph \\
    $S_{10}$ & IMS & Solo Fast Lap & $>$ 140 mph \\
    $S_{11}$ & IMS & Pylon Avoidance & $<$ 70 mph \\
    \hline
  \end{tabular}
  \caption{RACECAR Scenarios}
  \label{tab:drivingscenarios}
\end{table}

\subsection{Racing Scenarios}
Table~\ref{tab:drivingscenarios} shows the eleven different racing scenarios which are included in the RACECAR dataset. 
There are 6 scenarios $S_1 \cdots S_6$ from LVMS and 5 scenarios $S_7 \cdots S_{11}$ from the IMS track. 
The scenarios are further categorized on the basis of solo, multi-agent, slow-speed, and high-speed runs with speeds indicated in Table~\ref{tab:drivingscenarios}. 
Scenario $S_6$ is especially exciting since it contains several multi-agent runs between pairs of several teams at speeds of over 130mph. 
$S_{11}$ is a scenario from IMS with static obstacle avoidance. 
By spanning these eleven scenarios, the RACECAR dataset provides an interesting mix of solo laps, multi-agent laps, overtaking situations, high-accelerations, banked tracks, obstacle avoidance, and pit entry and exit at different speeds.

\section{RACECAR: Data Organization}\label{sec:dataorg}

Each team's autonomous racing stack used the Robot Operating System (ROS) middleware. The raw sensor data for each scenario was logged using a ROS2 bag format. 
ROS2 bags are a popular database storage format and a variety of tools have been written to allow one to store, process, replay, and visualize bag data.
Each data frame is composed of a serialized message and a UNIX timestamp.

\subsection{Data Processing and Synchronization}
ROS bag data has been preprocessed in several ways. 
We pruned the bag files such that large swaths of time spent idling in certain parts of the track are removed. 
The next step was converting all the GNSS data into a uniform Cartesian coordinate system across the entire RACECAR dataset. 
This is especially useful for multi-agent scenarios ($S_5, S_6$) where we have positions of both vehicles on track expressed in the same global Cartesian coordinate frame.
The two data sources from different teams running in the same session were synchronized using both the UNIX timestamps in the data, as well as reported GPS satellite time. 

\subsection{Ground Truth Annotations}
Currently, the data is not professionally annotated with bounding boxes, but we provide centimeter-level accurate positions of the vehicles in a common global coordinate frame to evaluate perception benchmarks.
The Real-Time Kinematic (RTK) GNSS accuracy is below 1-2 cm. Included in the GNSS data message is a standard deviation value that drops below 0.05 m when a RTK fix is acquired. 
Also included in each packet is a solution status which denotes the uncertainty of the RTK correction. Centimeter level accuracy for a racecar of footprint 4.918 m x 1.886m is very accurate and a reasonable margin of error.



\section{ROS2 to NuScenes}\label{sec:nuscenes}
\begin{figure}
  \centering 
  \includegraphics[width=\columnwidth]{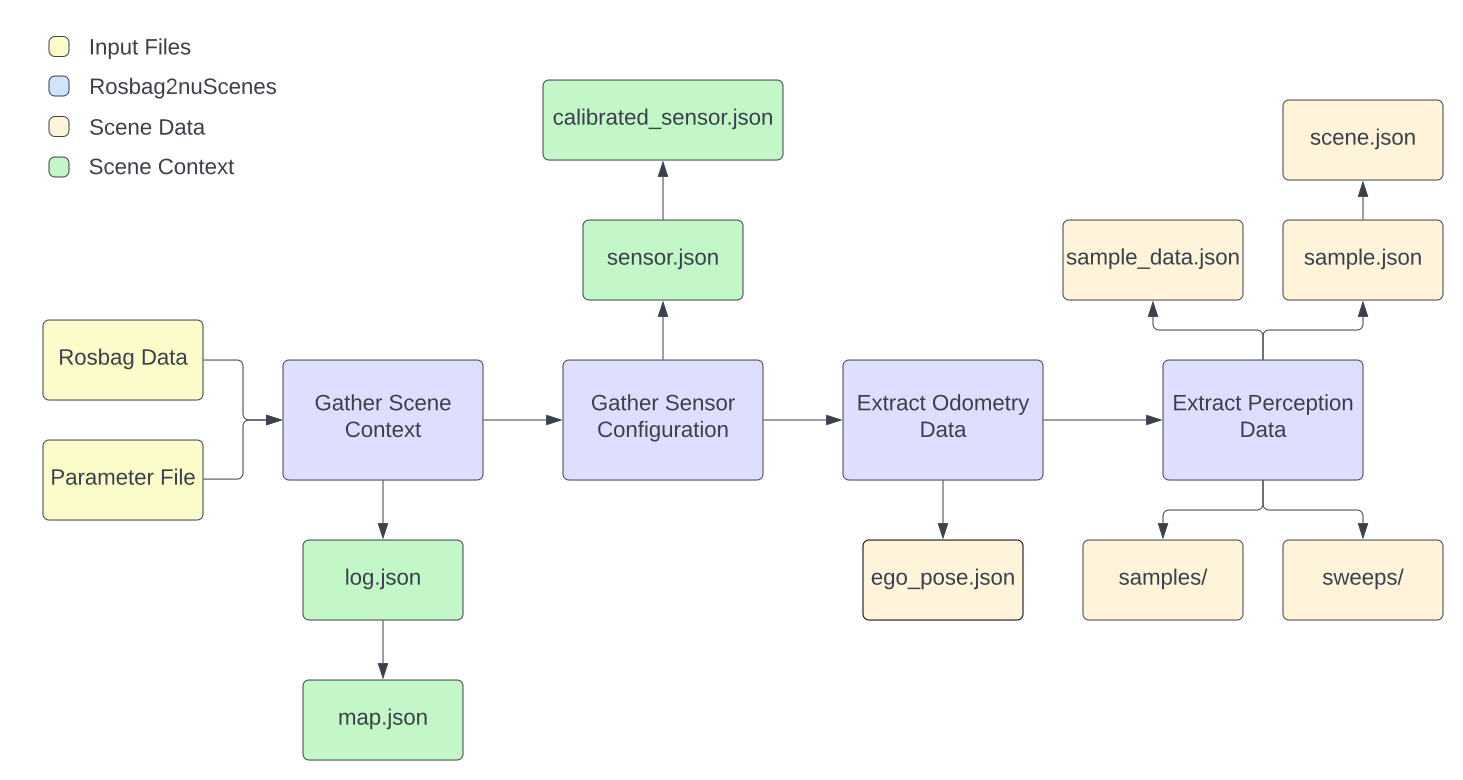}
  \caption{ROS2 bags to nuScenes conversion process}\label{fig:nuscenes_conversion}
\end{figure}

As stated earlier, one of the motivations behind releasing the RACECAR data is to present the data to as many researchers as possible.
Therefore, we also convert the entire dataset into the nuScenes format to improve the dataset's accessibility to the general self-driving community. 
\begin{figure*}
  \centering 
  \includegraphics[width=\textwidth]{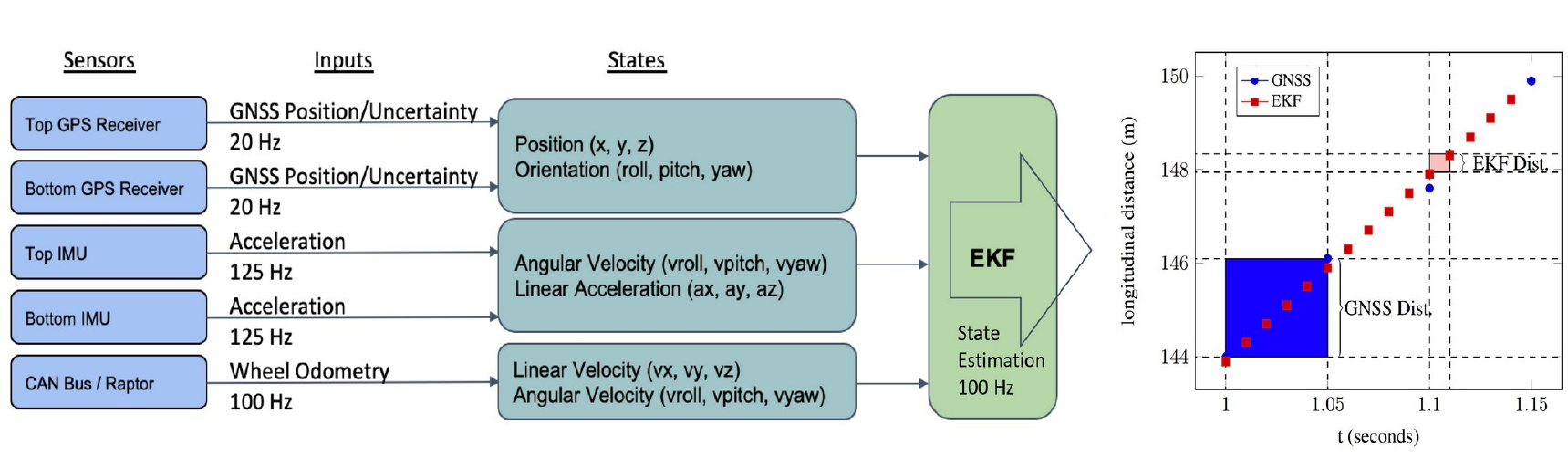}
  \caption{[Left] Localization Pipeline, sensors produce state measurements which are used as updates for an Extended Kalman Filter to produce a vehicle Pose estimate. [Right] The blue dots represent GNSS updates at 20 Hz. The red dots are EKF predictions at 100 Hz. At approximately 40 m/s, the car travels 2 meters before another GNSS update.}
  \label{fig:ekfpipeline}
\end{figure*}

In nuScenes format, data is organized in a tiered structure, starting with a scene at the highest level. 
A scene consists of a continuous stream of data for which a scenario took place, along with information on what occurred in the scene, the location of the scene, and when it was recorded. 
For our data, scenes correspond to scenarios (from Table~\ref{tab:drivingscenarios}).
The continuous stream of data present within a scene is used to create samples at fixed time intervals, containing all sensor measurements (e.g. LiDAR, Radar, or camera). 
All sensor data is recorded with a corresponding pose of the ego vehicle in an inertial frame and an extrinsic matrix to convert sensor readings to the inertial frame. 
The intrinsic calibration of camera sensors is also included to project between 3-dimensional coordinates and the image plane.  

The \texttt{rosbag2nuscenes} library was developed to convert the data originally stored in ROS2 bag files to the nuScenes format. 
An overview of the conversion process is shown in Figure \ref{fig:nuscenes_conversion}. 
Using the rosbag2 Python API, the conversion library reads through the database entries and extracts the necessary ROS2 messages, converting them to JSON files specific to the nuScenes schema.
The sampling rate is set at the default rate of 2 Hz, but is adjustable as laps at higher speeds warrant higher sampling rates than those at lower speeds. 
Scenes also last significantly longer than the twenty-second clips in the original nuScenes dataset.
\section{Autonomous Racing Benchmarks}\label{sec:benchmarks}

The RACECAR data contains full multi-modal sensor coverage of 11 exciting racing situations with full-scale autonomous racecars. 
Therefore, this dataset presents a unique, high-speed version of seminal problems in autonomous driving such as localization, object detection and tracking, and mapping. 
We present these three problems and demonstrate the applicability of the RACECAR dataset to establish new benchmarks for these problems within the context of high-speed autonomous racing. 
\subsection{Benchmark 1: Localization}\label{sec:localization}


High precision and low latency localization is a key challenge of autonomous racing~\cite{lingemann2005high}.
At a speed of 150 mph, the racecar travels up to 220 ft in one second.
Since all of the racing took place outdoors and under clear sky conditions, the localization methodology adopted by several teams was mainly based on a fusion of the two GNSS signals and their IMU units using an Extended Kalman Filter~\cite{EKF1963} as shown in Figure~\ref{fig:ekfpipeline}.
LiDAR-based~\cite {lidarloc} and camera-based~\cite {imageslam} localization are also possibilities and the RACECAR dataset will enable such a comparison.

\begin{figure}
  \centering 
  \includegraphics[width=1.0\columnwidth]{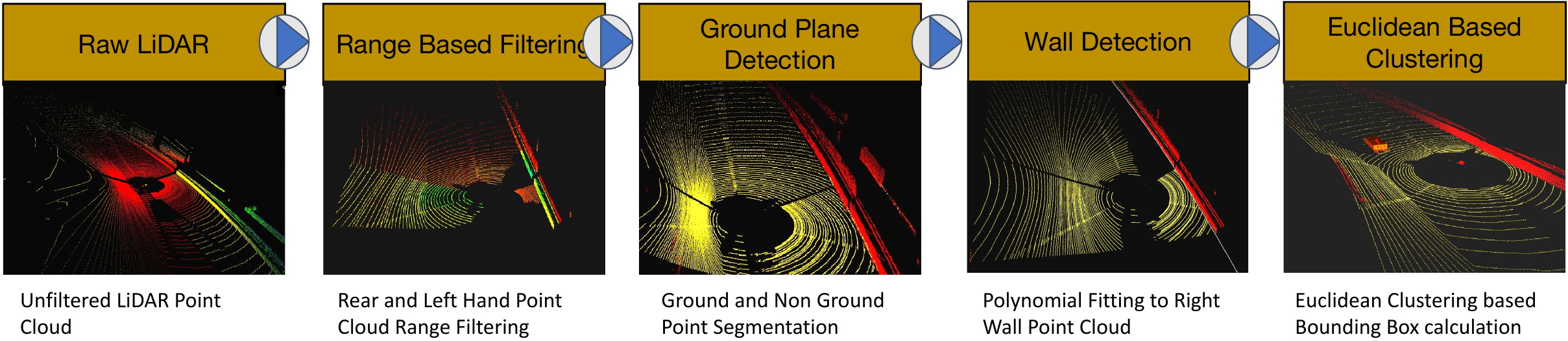}
  \caption{LiDAR Perception: Processing raw LiDAR point clouds using Range Based Filtering, Ground Plane Detection, Wall Detection, and Euclidean Based Clustering }
  \label{fig:perceptionpipeline}
\end{figure}

Using GNSS data alone results in a localization rate of only 20 Hz. 
While this may be fine for passenger autonomous driving, it is not adequate for autonomous racing. 
The AV-21 reports the rotation of the wheels at 100 Hz and accelerometer and gyroscope data at 125 Hz. This information is fused together with successive GPS readings and the pose estimate of the racecar $[x,y,z,\theta]$ is obtained at 100Hz. Here $x,y,z$ corresponds to the Cartesian location of the car in world frame, and $\theta$ is the vehicle heading. 

Figure~\ref{fig:ekfpipeline}[Right], shows the position estimates for the racecar traveling at 40 m/s (90 mph). 
A comparison between GNSS only and an EKF estimate is shown.
It can be seen that at this speed the racecar can travel up to 2m blindly before another estimate of position is received via the GNSS.
With an EKF running at a 5x faster rate, only 0.4m is traveled before another precise estimate is made resulting in more precise localization of the vehicle.


\subsection{Benchmark 2: Object Detection and Tracking}\label{sec:objdetect}

Long-range and robust detection and tracking of opponents on the track is of paramount importance.
At high speeds, overtaking another vehicle provides a very small window of time for the ego vehicle to react.
A false negative detection could result in a collision between the attacker and the defender.
Similarly, a false positive can cause erratic behavior by causing a vehicle to try and avoid a vehicle that is not there.
In this section, we present three example approaches for object detection and tracking using the LiDAR, Cameras, and Radar. 
Ideally, one would fuse all the detections together into one cohesive detection but we present these separately to showcase that the RACECAR data can enable object detection challenges for each sensing modality as well as challenges for fused detection. 

\begin{table}
\centering
\begin{tabular}[c]{ ||p{1cm}|p{1cm}|p{1cm}|p{1cm}|p{1cm}|p{1cm}||}
 \hline
 \multicolumn{6}{|c|}{Birds Eye View (BEV)} \\
 \hline
 Model   & Overall & 0-20m & 20-40m & 40-60m & 60m-Inf \\
 \hline
 PP AP & 83.02 & 99.72 & 98.73 & 93.08 & 54.27 \\
 VR AP & 75.82 & 97.05 & 90.39 & 83.60 & 40.71 \\
 \hline 
 \multicolumn{6}{|c|}{3D Bounding Box} \\
 \hline
 PP AP & 71.62 & 90.37 & 89.69 & 79.08 & 37.33 \\
 VR AP & 63.31 & 86.17 & 85.85 & 65.62 & 23.95 \\
 \hline
\end{tabular}
\caption{Baseline AV-21 LiDAR Detections trained on LVMS Multi-Agent Data, using both Birds Eye View (BEV) labels and 3D Bounding Boxes. PP: PointPillars, VR: VoxelRCNN, AP: Average Precision}
\label{tab:dnn_benchmark}
\end{table}

\subsubsection{LiDAR Based Object Detection and Tracking}

3D point clouds obtained from the LiDAR are quite dense and noisy for object detection. 
Figure~\ref{fig:perceptionpipeline} shows an Euclidean clustering based pipeline for object detection. 
This method involves downsampling the point cloud, region of interest based filtering, ground-plane segmentation, and then an euclidean distance based clustering algorithm.
This classical approach to object detection for point clouds is limited to a detection range of around 50 m.
Machine learning approaches to object detection are very popular but due to the requirement of annotated training data, exploration of auto-labeling methods, synthetic data generation, or hand labeling is necessary.
With the RACECAR data, we hope deep learning approaches to racing problems can be further explored.
We have provided results of a baseline implementation of both PointPillars \cite{pointpillars} and VoxelRCNN \cite{voxelrcnn}, shown in Table \ref{tab:dnn_benchmark}.
These models were trained on two multi-agent runs within the RACECAR dataset, with labels generated from the GNSS/IMU information provided.
Inference on point clouds using PointPillars resulted in maximum detections at distances up to 110 m. However, the average precision of detections dropped approximately 40$\%$ for distances over 60 m.
For racing speeds of over 100 mph, 60 m detections provide a second of reaction time, and for computationally expensive planning algorithms, every additional moment counts.
A benchmark challenge is to improve the detection range and reliability.



\subsubsection{Radar Based Object Detection and Tracking}


\begin{figure}
  \centering 
  \includegraphics[width=\columnwidth]{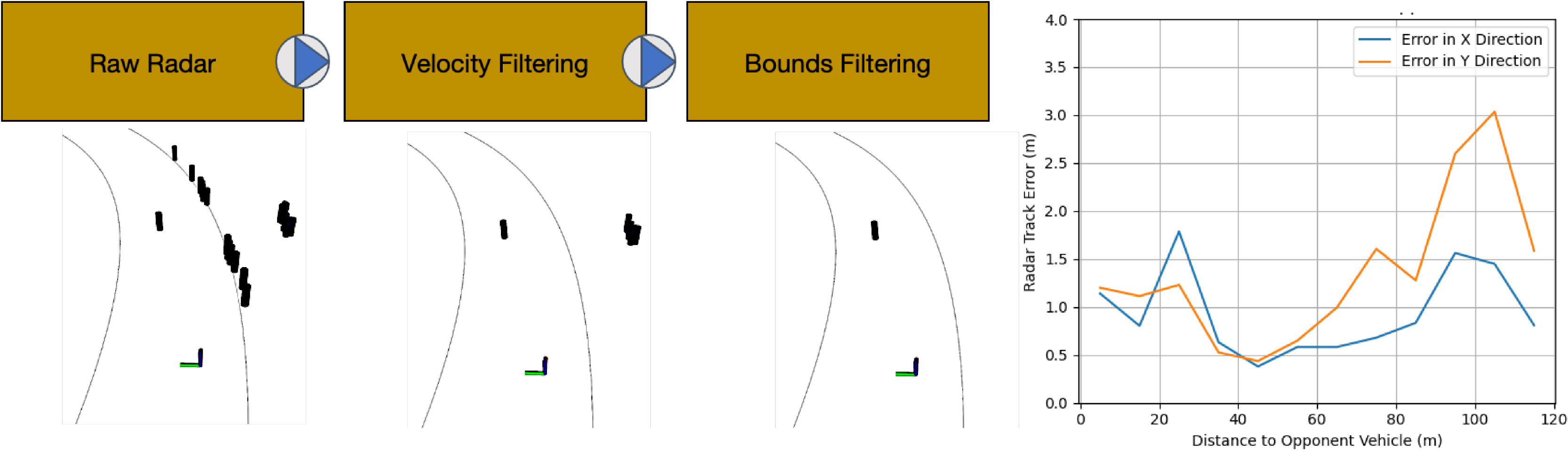}
  \caption{[Left] Radar Filtering based on Track Boundaries and Velocity. [Right] Tracking Error when Compared to Ground Truth for an overtake between TUM Motorsoprt and EuroRacing team.}
  \label{fig:radarerror}
\end{figure}

The front Electronically Scanning Radar (ESR) returns a list of tracked objects within its frame of view.
Each object tracked is packaged with data concerning its angle, distance, forward velocity, and lateral velocity all with respect to the radar.
Similar to the raw LiDAR data, at high speeds the Radar data is noisy, returning tracks for arbitrary points on the racetrack wall in addition to random objects nearby.
One strategy for removing undesirable objects is to filter by velocity, dynamically adjusting based on ego speed, as well as a region of interest filter similar to the LiDAR.
Figure~\ref{fig:radarerror} describes the error in both x and y directions between the Radar detected position and the ground truth position from a head-to-head racing scenario between TUM and EuroRacing.

\subsubsection{Camera Based Object Detection and Tracking}


Onboard the AV-21 are six color, global shutter cameras. 
To maximize detection range, the front-facing two cameras utilize a narrower camera lens, as this improves far-field resolution. 
The remaining four cameras use a wider field of view to provide $360^\circ$ coverage around the vehicle. 
Example images from the cameras can be seen in Figure~\ref{fig:overview}.

\noindent \textbf{Calibration:} As part of the data set, full camera intrinsics and extrinsics are provided. 
Instrinsics were obtained using an off-the-shelf camera calibration package available in the ROS 2 ecosystem. 
Extrinsics were obtained using laser rangefinders and surveying equipment, measuring the sensor locations with respect to a fixed point on the vehicle. 

\begin{figure}
  \centering 
  \includegraphics[width=\columnwidth]{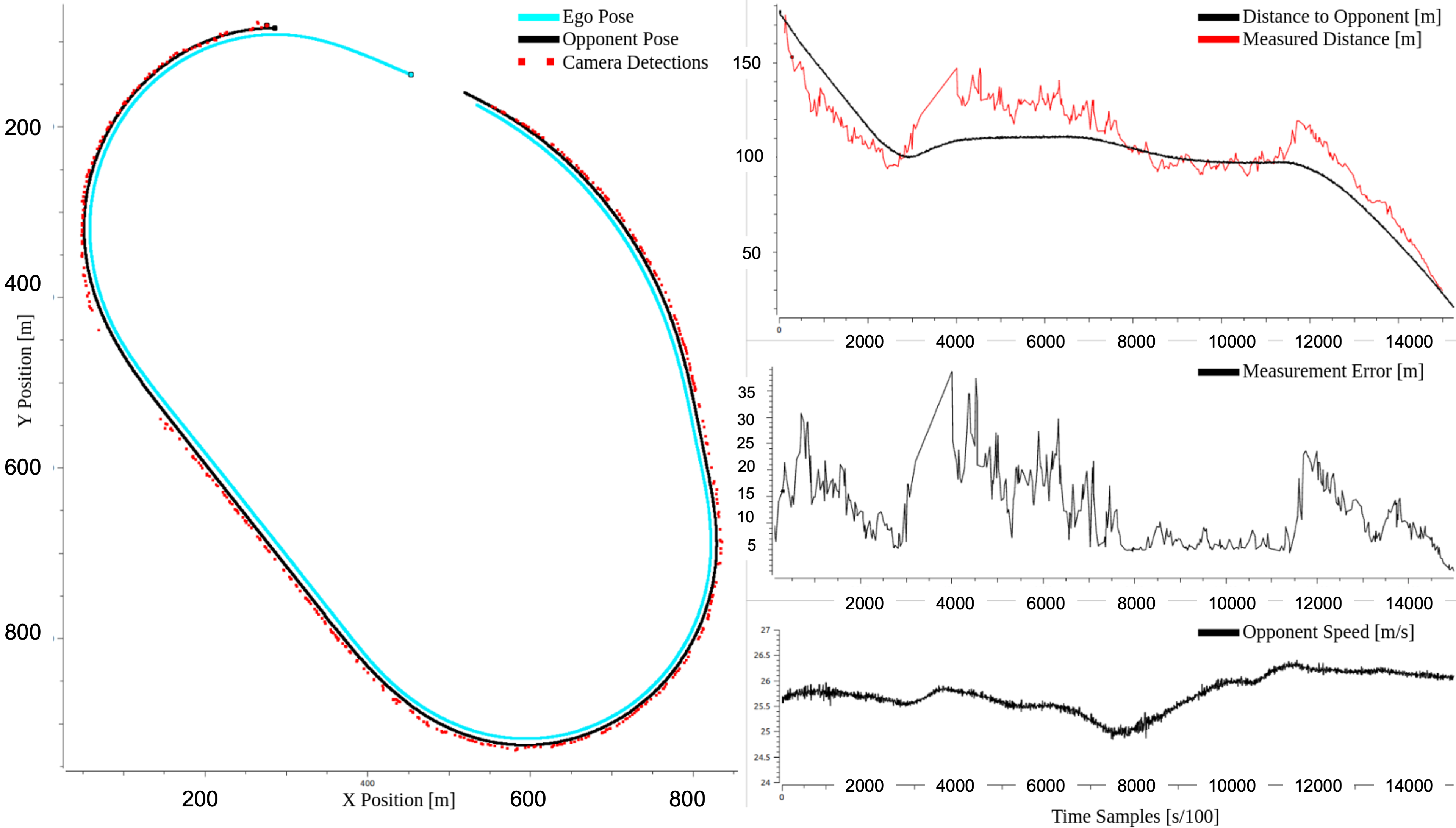}
  \caption{(Left) RTK-GPS ground truth position of the ego vehicle and opponent, overlaid with camera detections. (Top Right) Distance from ego to the opponent. (Middle Right) Error between measured and ground truth. (Bottom Right) Speed profile of opponent 
  }
  \label{fig:cam-gps-compare}
\end{figure}

\noindent \textbf{Camera Object Detection Methods:} Due to the higher resolution and higher frame rates, cameras have the potential to provide faster and longer-range detections of opponent vehicles. However, camera detections are inherently noisier and more difficult to localize in 3D, due to depth ambiguity and projection error.
A baseline utilizing YOLO v5~\cite{yolo_v5} was trained using a data set of other AV-21 vehicles. 
By exploiting the fact that the dimensions of the AV-21 are known, we can extend YOLO to report back real-time object depths by using a standard pinhole optics model. 
\begin{figure}
  \centering 
  \includegraphics[width=\columnwidth]{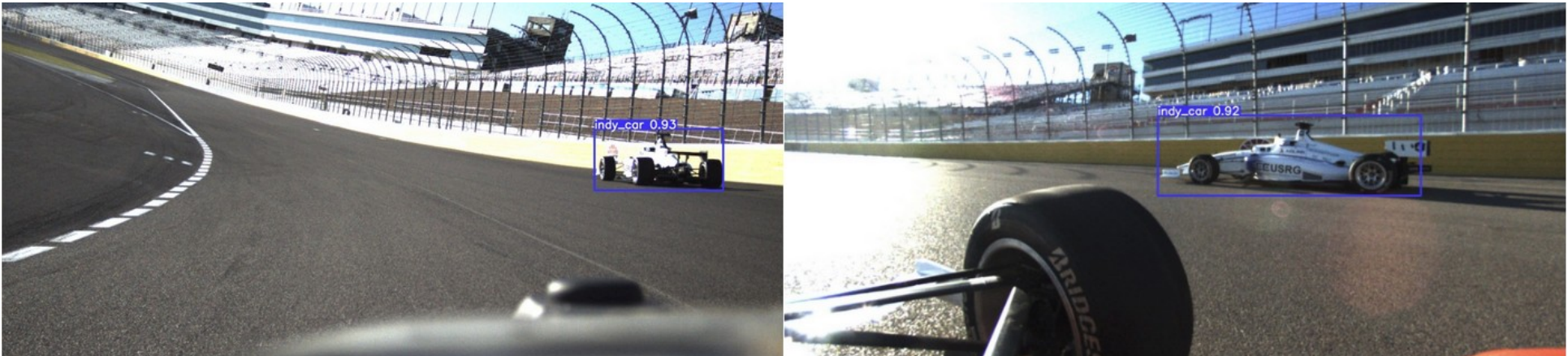}
  \caption{Detections from YOLO on several camera views during a passing maneuver between MIT-PITT-RW and KAIST.}
  \label{fig:cam-detection-collage}
\end{figure}
Detections from several camera views during a passing maneuver between MIT-PITT-RW and KAIST can be seen in Figure \ref{fig:cam-detection-collage}. 
The baseline was also compared to the GPS position of the opponent vehicle during a lap on the Las Vegas Motor Speedway track as shown in Figure~\ref{fig:cam-gps-compare}.

\subsection{Benchmark 3: Mapping}\label{sec:mapping}

\begin{figure}
    \centering
    \includegraphics[width=\columnwidth]{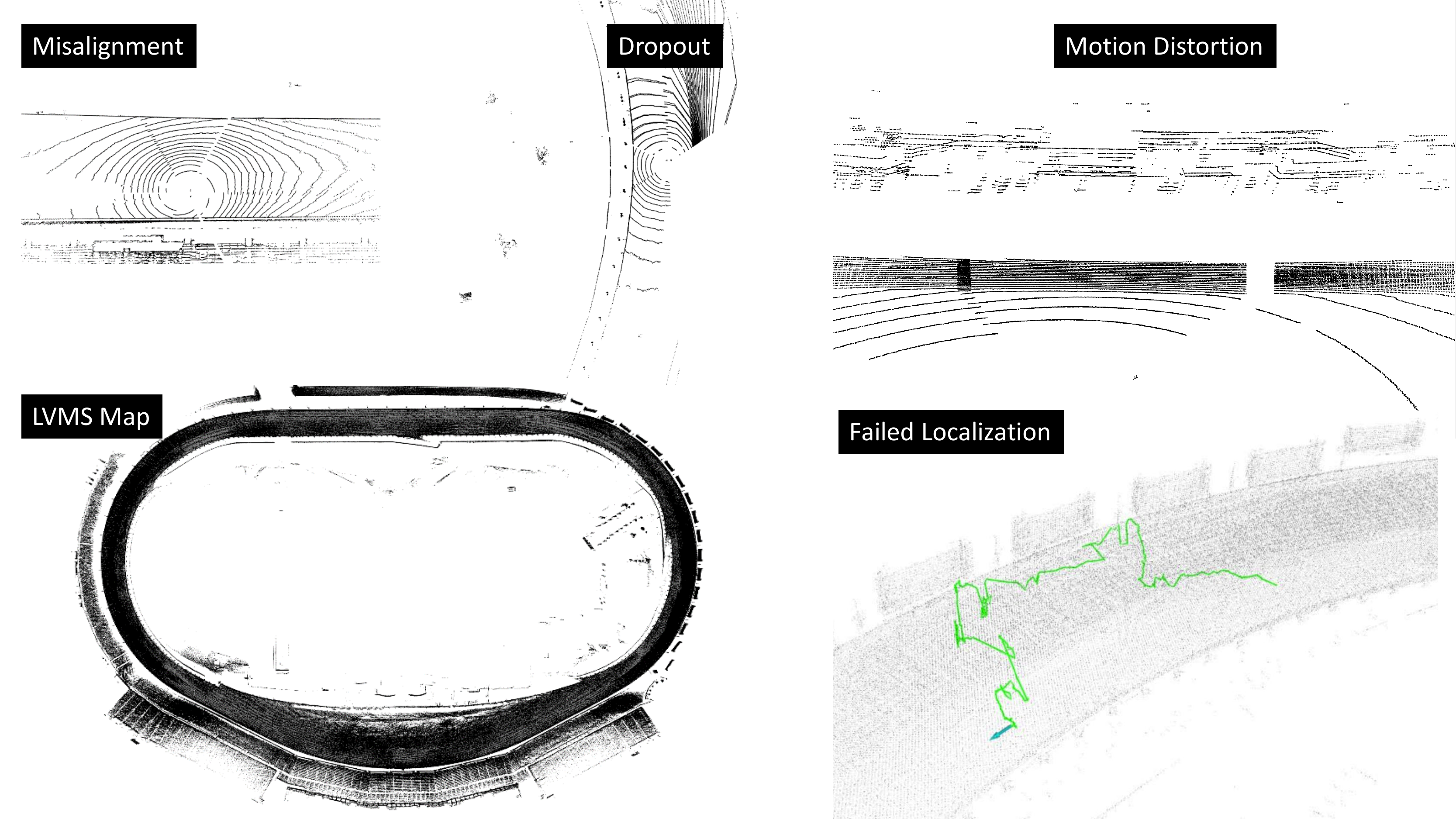}
    \caption{Clockwise from Top Left: Faulty calibration of the LiDARs. Dropout of the rear right LiDAR. Motion distortion due to high velocity.Failed localization on a map. SLAM generated 3D point cloud map for LVMS. }
    \label{fig:lidarfig}
\end{figure}

Relying on one method of localization can be risky, and in the case of having faulty or unreliable GNSS sensors, it can be useful to have an alternative solution. 
Implementing Simultaneous Localization and Mapping (SLAM) using 3D LiDAR point clouds or camera images for 2D visual SLAM, can provide another source of localization and an online method of mapping. 
Point cloud mapping aims to build a 3D point cloud map of an environment from sensor data that conveys 3D information about the surroundings of a perceiving agent, either directly like a LiDAR or indirectly as in the case of 2D visual SLAM. 
The RACECAR dataset contains highly interesting data for implementing LiDAR SLAM algorithms, however several key challenges remain. 
Since three LiDARs are used, precise calibration is crucial to allow good scan matching. Bad calibration can be seen in Figure~\ref{fig:lidarfig}.
High velocity leads to big jumps between the individual scans. This makes the SLAM more dependent on a good initial guess for the transformation of each frame.
High velocity also leads to strong motion blur. This needs to be compensated before matching the point clouds. 
Figure~\ref{fig:lidarfig} shows the effect of motion distortion on the tents behind the track barrier that are vertical in reality. Due to the blur, they appear sheared in the point cloud.
The banked turns make it impossible to assume a flat ground. 
Available SLAM algorithms tend to create spiral maps. A new handling of the ground surface has to be implemented.

\noindent \textbf{Map based Localization:}
To avoid these problems, map-based localization approaches can be realized. 
Online, the current scan can be matched with the offline-generated map. However, this is still a difficult task as the current scans have to be corrected first, and good state estimation is necessary to provide initial guesses for scan matching. 
Existing SLAM packages fail to provide a precise and robust localization output on the RACECAR dataset. 
Figure~\ref{fig:lidarfig} shows a failed SLAM attempt at the LVMS track. 
Robust SLAM at high speeds (100+ mph) is still a challenge for autonomous racing. 

\section{Conclusion}
The paper presents the RACECAR dataset for high-speed autonomous racing. 
Multi-modal sensor data has been collected, processed, and converted into ROS2 bag files and nuScenes format for wider accessibility. 
The data was collected from AV-21 autonomous Indy Light racecars during the 2021-22 Indy Autonomous Challenge held at the Indianapolis Motor Speedway and at the Las Vegas Motor Speedway tracks which witnesses overtaking at speeds of 170 mph. 
There are 27 racing sessions spanning 11 racing scenarios and over 6.5 hours of data that have been provided. 
We provide ground truth locations for the vehicle which enables use of the data for several benchmark problems in autonomous racing - localization, object detection and tracking, and mapping. Baseline algorithms are presented for each benchmark. The dataset, accompanying processing scripts, and the ROS2 to NuScenes conversion library are all open source.


{\small
\bibliographystyle{ieeetr}
\bibliography{egbib,mb}
}

\clearpage
\appendix
\begin{table*}
\centering
\begin{tabular}[c]{ ||p{3cm}|p{1.5cm}|p{1.5cm}|p{1.5cm}|p{1.5cm}||}
 \hline
 \textbf{Scenario (Teams)}  & \textbf{GNSS} & \textbf{LiDAR} & \textbf{RADAR} & \textbf{Camera}   \\
 \hline
 $S_{1} (C, M, P)$ & 51,486 & 58,375 & 9,924 & 157,146 \\
 \hline
 $S_{2} (C, M, K)$ & 79,748 & 60,020 & 1,148,205 & 301,065 \\
 \hline
 $S_{3} (M, E)$ & 52,194 & 44,855 & 31,444 & 387,756 \\
 \hline
 $S_{4} (T, E)$ & 23,852 & 23,722 & 15,725 & 0 \\
 \hline
 $S_{5} (C, M, P, T, E, K)$ & 129,914 & 118,156 & 23,525 & 122,204 \\
 \hline
 $S_{6} (T, E, P)$ & 52,504 & 67,253 & 35,023 & 0\\
 \hline
 $S_{7} (C, K)$ & 16,596 & 8,025 & 0 & 0 \\
 \hline
 $S_{8} (C, K)$ & 22,140 & 0 & 0 & 270,648 \\
 \hline
 $S_{9} (T, E)$ & 44,510 & 40,569 & 23,129 & 0\\
 \hline
 $S_{10} (P)$ & 13,153 & 12,476 & 0 & 0 \\
 \hline
 $S_{11} (T, K)$ & 22,594 & 29,746 & 6,440 & 278,009 \\
 \hline
\end{tabular}
\caption{Dataset Content: Raw ROS2 message frame counts for each sensor}
\label{tab:Dataset}
\end{table*}

\section{Technical Appendix}

\subsection{Complete Sensor Specifications}

\begin{figure*}
    \centering
    \includegraphics[width=0.85\linewidth]{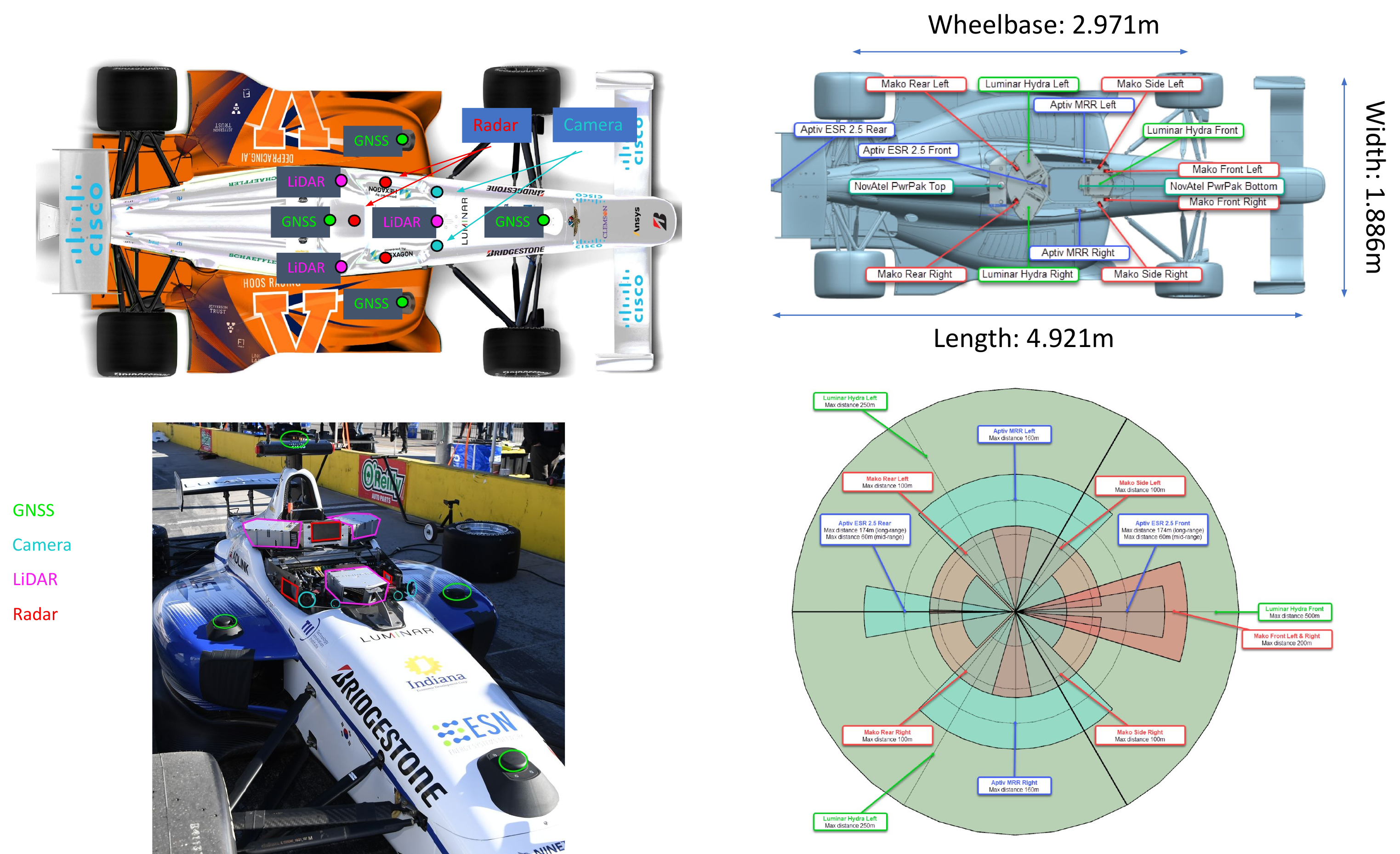}
    \caption{Top Left: Rendering of the racecar with visible sensors highlighted. Top Right: Vehicle diagram showcasing the length, width, and wheelbase of the vehicle. Bottom Left: Inside view of cockpit with various sensors highlighted. Bottom Right: Sensor ranges for LiDAR, Radar, and Cameras are overlayed on top of the vehicle.}\label{fig:sensorlayout}
\end{figure*}

The data set was captured from a suite of sensors chosen to help teams run a full autonomous driving stack.
The racecar was designed with the entire driver cockpit removed and replaced with a computer, drive-by-wire system, and sensors.

\subsubsection{LiDARs}

  
    The car is outfitted with three Luminar H3 Solid State LiDARs. 
  The LiDAR's output raw 3D point clouds comprised of x, y, z coordinates, and intensity for each laser scan. 
  The rate at which these messages are published is dependent on the density of the point cloud, with more laser scans producing a higher density point cloud but a lower frame rate. 
  The scan pattern of the LiDARs can also be configured in Uniform, Trapezoidal, or Gaussian distributions allowing higher point cloud density in focused areas. 
  As seen in the bottom right quadrant of Figure \ref{fig:sensorlayout}, the three LiDAR's field of view covers $360 \deg$, with a detection range of 250 m.

\subsubsection{Global Navigation Satellite System (GNSS)}

    GNSS and Inertial Measurement Unit (IMU) information were collected using two Novatel Pwrpak 7d Receivers and was the primary source of localization for the vehicles. 
There exist two receivers for the purpose of redundancy, and as they are stacked on each other within the vehicle they are referred to as the top and bottom receivers.
The receiver uses two symmetrically placed antennas on the AV21 seen in the top left of Figure \ref{fig:sensorlayout}.
Both receivers were configured with a Real Time Kinematic (RTK) system to provide positioning information with up to centimeter level accuracy. 
The onboard IMUs provide angular velocity and acceleration data, which help drive an Inertial Navigation System (INS) to dead reckon position, velocity, and heading.
The raw data provided here includes latitude, longitude, altitude, and their respective standard deviations from each receiver.
Also included are the acceleration and angular velocity collected from each IMU.
\subsubsection{Cameras}

    Six Allied Vision Mak G319C Cameras were installed on the vehicle.
The cameras resolution, frame rate, focal length, and optical center can be configured, and output raw uncompressed images.
These are uncompressed images and require further processing for translation into usable point clouds. 
The cameras possess functionality for Precision Time Protocol and allow device synchronization in the order of microseconds.
\subsubsection{Radar}


    The AV-21 has both long range and medium range Aptiv RADAR devices installed. 
An electronically scanning radar is placed for long range detection on the front of the vehicle, and two mid range radars are placed on the side of the vehicle.
Raw radar data is interfaced through the CANBus and after processing from a driver, provides detected vehicle speed, a covariance matrix, and relative velocity of detected objects.
The two radars range and field ov view is described in Table \ref{tab:Specifications} and can be seen in Figure \ref{fig:sensorlayout}.
\subsubsection{Drive-by-wire}
    The car was outfitted with a New Eagle Raptor drive-by-wire system to provide an electronic control interface for the vehicle actuators. 
The computing stack on the onboard computer provided steering angle, throttle input, brake input, and gear shifting commands to the Raptor interface. 
Battery voltage, engine temperature, engine RPM, current gear, and the wheel speed of the vehicle were all consistently monitored and assisted in control of the vehicle.
\subsubsection{Communication}
Each car used a Cisco Ultra Reliable Wide Band radio to connect to a track-wide mesh network. 
This network provided an internet connection for GNSS RTK corrections, as well as connection to a basestation computer used for live observation of the vehicle telemetry. 
The AV21 also communicated with race control, racetrack supervisors providing command flags indicating when cars should stop, slow down, or return to the pits.
Race control sent flags to the AV21s using MyLaps, a popular sports timing system which also provided vehicle telemetry to race control.
\subsubsection{Onboard Computing}
    Onboard computing of the car was handled by an ADLINK AVA-3501. 
The computer posseessed a Intel Xeon processor, 64 GB of RAM, 3 TB of storage, and a Quadro RTX 8000 GPU.
The computer was intended as a platform to run every component of the self driving computing stack, including any potential machine learning methodologies.
Each computer was setup with Ubuntu 20.04 and ran ROS2 to interface with each component of the vehicle. 
All the sensors, as well as the drive-by-wire system used ROS2 drivers to provide a consistent common interface.
\subsubsection{Chassis}

The vehicle chassis was manufactured by Dallara. 
The car resembles an AV-21 Indy Lights race car, but has been retrofitted to accomodate the various sensors and computing platforms described. 
All of the cockpit and safety features for a human driver were removed and replaced with a platform to house all the electronic components.

\subsection{Extended Kalman Filter}
Formulating and processing the EKF algorithm is a well known process, to estimate the full 3D (6 Degrees of Freedom) pose and velocity of a robot over time.

The first step is to represent one's process using a nonlinear dynamic system.

\begin{equation}
    \pmb{x}_{k} = f(x_{k-1}) + \pmb{w}_{k-1}
\end{equation}

$\pmb{x}_{k}$ represents the robot's current 3D pose at time k, $f$ is the state transition function, and $\pmb{w}_{k-1}$ is added noise towards the process.
The vector $x$ contains the robot's 3D pose, 3D orientation, and their derivatives.

\begin{equation}
    \pmb{\hat{x}}_{k} = f(\pmb{x}_{k-1})
\end{equation}

\begin{equation}
    \pmb{\hat{P}}_{k} = \pmb{FP}_{k-1}\pmb{F}^{T} + \pmb{Q}
\end{equation}

The algorithm begins with a prediction of the future state, using the current state estimate, the transition function, and estimated covariance. 
$\pmb{P}$ is the estimated covariance, $\pmb{F}$ is the Jacobian of the non linear transition function, and $\pmb{Q}$ is the process noise covariance that perturbs the system. The estimated covariance can be initialized to some reasonable value, but the process noise covariance should represent the uncertainty between the real life system model, and the approximate predictions made by the transition function. Tuning these values carefully requires experimentation and iteration, and should be kept at a relatively small value initially.

\begin{equation}
    \pmb{z}_{k} = h(\pmb{x}_{k}) + \pmb{v}_{k}
\end{equation}

Sensor measurements $\pmb{z}_k$ are recieved  at time $k$, $h$ is a sensor model that correctly maps measurement inputs into a state space vector, and $\pmb{v}_k$ is the measurement noise. These sensor measurements can be any state measurement from the IMU, wheels, GNSS measurements, or LiDAR laser scans. The measurement noise can be approximated from testing the hardware directly.

\begin{equation}
    \pmb{K} = \pmb{\hat{P}}_{k}\pmb{H}^{T}(\pmb{H\hat{P}}_{k}\pmb{H}^{T} + \pmb{R})^{-1}
\end{equation}

\begin{equation}
    \pmb{x}_{k} = \pmb{\hat{x}}_{k} + \pmb{K(z - H\hat{x}}_{k})
\end{equation}

\begin{equation}
    \pmb{P}_{k} = (\pmb{I - KH})\pmb{\hat{P}}_{k}(\pmb{I - KH})^{T} + \pmb{KRK}^{T}
\end{equation}

After the prediction step and a measurement update is received, the estimated state and it's associated covariance is updated in a correction step.
$\pmb{K}$ represents the Kalman gain, which is calculated from the measurement covariance $\pmb{R}$ and the estimated covariance $\pmb{P}$.
This gain is used to update the state vector and it's covariance matrix. After the correction step, another prediction step is made, and the process repeats.\

\begin{table}
    \centering
    \begin{tabular}[c]{ ||p{5cm}|p{4cm}||}
     \hline
     \multicolumn{2}{|c|}{AV21 Specifications} \\
     \hline
     \textbf{GNSS}   & \textbf{PwrPak7 E1}   \\
     \hline
     GNSS Positional Accuracy (L1) & 1.5m \\
     GNSS Positional Accuracy (RTK) & 1cm + 1ppm \\
     GNSS Data Rate & 0-20 Hz \\
     IMU Data Rate & 125 Hz\\
     \hline
     \textbf{LiDAR}   & \textbf{H3 Prototype}  \\
     \hline
     Range (10\% Reflectivity) & 250m \\
     Horizontal Field of View & 120\textdegree \\
     Configurable Vertical Field of View & 0-30\textdegree \\
     Frames per Second & 10-30 \\
     \hline
     \textbf{Front Radar}   & \textbf{Front Facing Aptiv ESR}  \\
     \hline
     Long Range Detection & 174m \\
     Mid Range Detection & 60m \\
     Long Range FOV & $\pm10$ deg \\
     Mid Range FOV & $\pm45$ deg\\
     Update Rate & $20$ Hz \\
     \hline
     \textbf{Side Radar}   & \textbf{Side Facing Aptiv MRR}  \\
     \hline
     Long Range Detection & 160m \\
     Mid Range Detection & 40m \\
     Horizontal Field of View & $\>90$ deg \\
     Vertical Field of View & $5$ deg \\
     \hline
     \textbf{Camera}   & \textbf{Mako G-319}  \\
     \hline
     Max Resolution & 2064 x 1544 \\
     Max frame rate at full resolution & 37.6 fps \\
     Spectral range & 300-1100 nm \\
     \hline
     \textbf{Chassis}   & \textbf{Dallara AV-21}  \\
     \hline
     Overall Length & 192 in / 4876 mm \\
     Overall Width & 76 in / 1930 mm \\
     Overall Height & 45.5 in / 1156.5 mm \\
     Wheelbase & 117 in / 2971 mm \\
     Weight & 1600 lbs / 726 kgs \\
     \hline
    \end{tabular}
    \caption{AV21 - Sensor Specifications}
    \label{tab:Specifications}
    \end{table}

\end{document}